# Matrix Domination: Convergence of a Genetic Algorithm Metaheuristic with the Wisdom of Crowds to Solve the NP-Complete Problem

Shane Storm Strachan
ssstra01@louisville.edu

***Abstract*** This research explores the application of a genetic algorithm metaheuristic enriched by the wisdom of crowds in order to address the NP-Complete matrix domination problem (TMDP) which is itself a constraint on related problems applied in graphs. Matrix domination involves accurately placing a subset of cells, referred to as dominators, within a matrix with the goal of their dominating the remainder of the cells. This research integrates the exploratory nature of a genetic algorithm with the wisdom of crowds to find more optimal solutions with user-defined parameters to work within computational complexity considerations and gauge performance mainly with a fitness evaluation function and a constraining function to combat the stochastic nature of genetic algorithms. With this, I propose a novel approach to TMDP with a genetic algorithm that incorporates the wisdom of crowds, emphasizing collective decision-making in the selection process, and by exploring concepts of matrix permutations and their relevance in finding optimal solutions. Results demonstrate the potential of this convergence to generate efficient solutions, optimizing the trade-off between the number of dominators and their strategic placements within the matrices while efficiently ensuring consistent and complete matrix domination.

## I.   INTRODUCTION

Matrix Domination is an NP-Complete problem introduced more than forty years ago[1] and has deep relations to concepts of graph edge traversal, and specifically the concept of the domination of a graph. In graph theory, dominating sets focus on vertices and their neighbors, and indeed matrix domination considers elements in a matrix and their relation to its other elements where each matrix cell corresponds to a vertex in a graph.

The goal is to place the smallest number of dominators in the matrix such that all cells are dominated, and a cell is considered dominated if it either a) contains a dominator or b) is orthogonally adjacent to a cell containing a dominator, which represents a complex relationship even in some of the more limited (smaller) contexts. Matrix domination has practical application in such areas as network design, logistics, surveillance, and resource allocation, among others. Here, approaching it in its "theoretical" nature, there is only consideration of dominators and the dominated (rather than "tiers" of domination, which would add much further complexity). Due to its NP-completeness, dominating large or complex matrices is computationally demanding due to its complexities and so should require the employment of a metaheuristic or approximation algorithm.

For clarity, the matrix domination problem (henceforth TMDP), differs from simply dealing with permutation matrices. A permutation matrix is a square binary matrix that has exactly one entry of 1 in each row and each column and 0s elsewhere with the purpose of performing a permutation of rows or columns in matrix multiplication. They are thus a very specific case of binary matrices with a very structured form. While TMDP involves a binary matrix (here 1s and 0s and represented visually as blue and black cells, respectively), it is instead a "search" for a subset of entries (the aforementioned dominators) that dominate other entries within its spheres of influence.

The goal of TMDP is to have every 0 in the matrix dominated, without violating established constraints (most importantly, that there are no redundant dominators). The problem is more complex than permutation matrices because it is not just about having a single 1 in each row and column, but rather it is about the *influence* a dominator has over the entries of the matrix, which introduces concepts like strategic value, risk assessment, and the distance factor as defined by the matrices' constructions.

The problem can be explicitly stated as follows:

Let there be an *n x n* matrix *M* with entries from { 0, 1 }, and a positive integer *K*. Is there a set of *K*, or fewer, non-zero entries in *M* that dominate all others? In other words, can *s* subset $C \subseteq \{ 1, 2, ... , n \} \times \{ 1, 2, ... , n \}$ with $|C| \leq K$ such that $M_{ij} = 1$ for all $(i, j) \in C$ and such that, whenever $M_{ij} = 1$, there exists an $(i', j') \in C$ for which either $i = i'$ or $j = j'$?[4]

As shown, the question asks if there is a certain kind of subset within a matrix, composed of pairs of indices from the matrix (representing the rows and columns). The size of *C* is constrained by *K* (meaning it has *K* or fewer elements) and domination means that for every non-zero entry in the matrix (*M* = 1), there is an entry in *C* such that they are either in the same row (*i* = *i'*) or column (*j* = *j'*). Importantly, TMDP remains NP-complete even if the matrix M is in a specific form, including very small, upper triangular matrices (where all entries below the main diagonal are zero).[*ibid*]

## II.   Prior Work

In this research, I propose a hybrid model that begins by integrating a genetic algorithm metaheuristic. This necessitates functions to establish an initial "population", to evaluate the population's fitness, to create selections, crossovers, and mutations. With an added layer of decision-making influencing the choices of the selection steps, the wisdom of crowds is implemented during selection and mutation phases, allowing for collective influence on the evolutionary trajectory of the solutions. The rest of this section will review and analyze the nature of NP-Completeness, the concept of domination in graph theory, the concept of a genetic algorithm, the concept of wisdom of crowds, and the nature of using those two methods to approach an NP-Complete problem related to domination.

As mentioned, TMDP is classified as an NP-complete problem. This classification implies that there is no known algorithm that can solve all instances of this problem in polynomial time. NP-Complete problems are fundamental to computational complexity and the NP stands for nondeterministic polynomial (time). In this context, an NP problem is one where a solution, if given, can be *verified* in polynomial time but actually *finding* the solution itself is not nearly as simple, straightforward, or efficient.[2] NP-Complete problems are known for their intrinsic difficulty as well as their interconnectedness with each other: a problem is classified as NP-Complete if it is both in NP and as hard as any problem in NP.[4]

The Cook-Levin Theorem was the first to demonstrate an NP-Complete problem. Their theorem laid the foundation for the theory of NP-Completeness, suggesting that if any NP-Complete problem can be solved quickly (in polynomial time), then every NP-Complete problem can also be solved quickly.[4] Furthermore, the study of NP-Complete problems involves the concept of reducibility, which is a way of comparing the complexity of different problem (e.g. Problem A is said to be reducible to a Problem B if there exists an efficient method to transform instances of A into instances of B such that the solutions for A can be used to solve B).

NP-Complete problems are generally quite simple to state and oftentimes even simple to plan out approaches to find solutions, however their difficulty lies in their rates of increase with relevance to the size of the problem. Their complexity grows exponentially. Related to NP-Completeness is the concept of NP-Hardness: A problem is NP-Hard if it is at least as hard as the hardest problems in NP, but it might not necessarily be in NP itself.[ibid] NP-Complete problems, by definition, are NP-Hard but not only so.

Delving into more detail on domination in graph theory, domination is an important concept that focuses on sets of vertices within a graph that exert influence or control over the areas of the structure. A set of vertices in a graph is a dominating set if every vertex not in the set is adjacent to at least one vertex.[16] This concept is crucial in applications such as network design, social network analysis, and resource allocation problems, where there is a need for the assurance that certain nodes (that could represent anything from a building to a set of servers to a person, etc.) can effectively serve or control other nodes in the network.

Total domination in graph theory is a variation of this concept, where the set must dominate itself as well as the rest of the graph.[12] This means that for every vertex there must be another vertex to which it is adjacent. Total domination is a strict requirement and finds utility in scenarios where self-reliance or internal regulation within a network is necessary. As an example, in a surveillance network each camera should not only monitor unsurveilled areas but also need to be surveilled themselves. Edge domination introduces another layer to this concept, where the focus shifts from vertices to edges. An edge dominating set in a graph is a set of edges such that every edge not in the set is adjacent to at least one edge in the set.

The minimization of dominating sets, total dominating sets, and edge dominating sets are important and often correspond to cost minimization in practical applications.[11] However, because finding the dominating sets is NP-complete, the task is computationally challenging for large problem sets.[4]

Indeed, TMDP can be visualized as taking place on a uniform grid where each cell can take one of two identities: dominator and dominated. The challenge is to find the minimum number of cells that need to be dominated such that every cell in the grid is either dominated or orthogonally adjacent to a dominated cell.

The complexity of TMDP lies in the vast number of possible combinations of cells that could form a successful dominating set, making exhaustive search methods impractical for large problem sets (in this case, large matrices). The problem's NP-completeness means that there is no known algorithm capable of solving all instances of the problem efficiently. To tackle the domination problems, heuristics and approximation algorithms have been developed. These algorithms aim to provide solutions, though not necessarily optimal ones, within a reasonable amount of time and computational and energy expenditures. For instance, a greedy algorithm would iteratively select cells that dominate the maximum number of undominated cells and can be used, though their weaknesses are quite clear for the reasons just stated. While such methods do not guarantee the most optimal solution with the smallest dominating set, they are often enough to find a tolerable solution set.

Another interesting aspect of TMDP is its relationship with other graph theoretical problems. For example, it is closely related to the maximum independent set problem and the vertex cover problem, both of which are also NP-complete.[4] This could provide some insight into the underlying structure of NP-complete problems. Their NP-complete status underscores the complexities inherent in their "real world" application, from network design and analysis to cryptography to pure mathematics.

With the concepts of NP-Completeness and graph domination in mind, I will now begin to discuss my approach to the NP-Complete TMDP using a genetic algorithm and the wisdom of crowds. Genetic algorithms are inspired by the processes of evolutionary biology and specifically the idea of natural selection. These algorithms apply Darwinian principles and the concepts of reproduction and survival to search out spaces and find optimal solutions (thus, a metaheuristic). A genetic algorithm operates by creating a population of candidate solutions to a given problem and these solutions undergo processes akin to biological competition and mutation.

By iteratively selecting for the "fittest individuals", the algorithm evolves the population toward an optimal solution. Unlike traditional optimization methods, which struggle in complex, multi-modal landscapes, genetic algorithms instead are efficient at navigating complex spaces. This is done by maintaining a diverse population of solutions, avoiding the pitfall of local optima. Their stochastic nature, however, means that they do not always produce the same results on each run even with the same parameters set, offering a range of optimal solutions rather than a single, deterministic optimum.

The algorithms typically share several key characteristics: selection, crossover, and mutation. Selection involves choosing the fittest individuals to reproduce, based on a fitness evaluation. Crossover is the process of combining two solutions to produce a new one, ensuring the mixing of "genetic material". Mutation, importantly, introduces random change to individuals, ensuring the maintenance of diversity in the population and enabling further exploration of the search space. These steps are then repeated over a set number of generations, with the algorithm converging on its most optimal solution given what its population is.

The concept of the "wisdom of crowds" is based on the argument that the collective judgment of humans often leads to more accurate and effective decisions than the conclusions of individual experts.[17] This idea is fundamentally rooted in the concepts of diversity of opinion and decentralization; the elusive "good outcome" of a mob mentality. The underlying principle is that in a large group, even if individuals are not exceptionally knowledgeable of the topic, the *aggregation* of their decisions can get to a sort of "better" outcome.

However, as hinted at, the effectiveness of crowd wisdom is contingent on certain assumed conditions: a diversity of opinions, the independent agency of its members, the decentralization of language and knowledge, and an effective method for aggregating such opinions. When these conditions are not met, the crowd remains prone to collective errors and biases.

## III.  PROPOSED APPROACH

Taking this into consideration, integrating the concept of the wisdom of crowds into genetic algorithms presents a novel approach to enhance solutions to TMDP. In a genetic algorithm, a population of solutions evolves over time through processes analogous to biological evolution., and by incorporating the wisdom of crowds, this process can be further improved. Their integration can be conceptualized as many populations within the algorithm, each evolving independently. These "wise" populations are representative of diverse approaches to the problem, ensuring that the search space is explored more thoroughly and reducing the risk of the algorithm converging prematurely on a less than ideal solution set.

The mutation and crossover, as mentioned, introduce variations in how dominators are placed within the constraints of one per row, one per column. This variation allows the algorithm to explore different configurations that might score differently according to the fitness evaluation function. Through generations, the genetic algorithm explores various configurations, guided by the fitness function. It progressively moves towards solutions that not only satisfy the basic constraints but also searches for more optima.

Attempting to properly position the dominators in a way that maximizes control over certain strategic points in the matrix it must be considered that two matrices might both satisfy the one dominator per row, one per column rule, but one might cover more strategic points effectively. The genetic algorithm helps find such optimal configurations. With TMDP, better solutions are those that satisfy the basic constraint of one dominator per row, one per column while *also* excelling based on additional criteria defined by the fitness evaluation function.

In the context of TMDP where the goal is to have exactly one dominator per row and column, the concept of a "better" solution is at first ambiguous and necessitates some thoughts on how to approach it. The exploratory nature of using a genetic algorithm metaheuristic allows for the conception of defining "better" for matrix domination in several important contexts: the exploration and learning that occurs through exploration and by applying constraints and penalties for attempting to break said constraints and the algorithm learning that this is undesirable.

The fundamental requirement for the problem is that only one dominator can exist in each row and column (diagonal adjacency is not relevant here). Any solution that meets this constraint is a valid solution, but not necessarily an optimal one. This is why the fitness evaluation function is critical: by assessing each solution not just on meeting the basic constraints, but also on how well it meets these additional criteria, the fitness function quantifies the "goodness" of a solution, allowing the algorithm to distinguish between merely valid solutions and more optimal ones.

Briefly going into more depth on the fitness evaluation, this function when called imposes a penalty if certain conditions are not met in each row and column of the solution matrix: here let *m* and *n* be the dimensions of the solution matrix *S*. This penalty formula imposes a numerically-defined punishment if certain conditions are not met in each row and column of the solution matrix, thereby discouraging certain characteristics in the solutions.

Let *P* here be the penalty value applied to *S*. The formula consists of two double summations, since it needs to sum over a two-dimensional matrix *S* with dimensions *m* by *n*. $S_{ij}$ represents an element of this matrix, where *i* is the row index and *j* is the column index. The Iverson brackets around $S_{ij} \neq 1$ means the expression inside the brackets is counted as 1 if the condition is true and 0 if the condition is false. The first double summation counts the number of elements in each row that are not equal to 1, summing this count across all rows. The second double summation counts the number of elements in each column that are not equal to 1, summing this count across all columns.

$$P = 10 \cdot \left( \sum_{i=1}^{m} \left[ \sum_{j=1}^{n} S_{ij} \neq 1 \right] + \sum_{j=1}^{n} \left[ \sum_{i=1}^{m} S_{ij} \neq 1 \right] \right)$$

These two counts are then added together. This means that the penalty is increased both for each not-1 element in every row and in every column, effectively penalizing the solution based on the combined count of non-1 elements across its entirety (not following the constraints of TMDP). Finally, the entire sum is multiplied by 10, scaling the penalty value.

In Solution matrix *S*, then, the penalty *P* increases with the number of elements in *S* that are not equal to 1. This is due to the value 1 indicating a dominator, whose successful positioning is paramount. The use of the penalty in the fitness function is to guide the genetic algorithm towards a solution where the elements of the matrix *S* include dominators to ensure complete matrix domination. Using this with the influence formula in the function, fitness is then defined as *I - P*, where *I* represents the influence, and minus P here is the subtraction of the penalty value from its influence value.

      Finally, it is important to reiterate the importance of stochasticity for genetic algorithms: this means they produce varied results. This is especially true in larger or more complex problem spaces, which is why constraining functions are necessary in addition to the fitness evaluation. The constraining function implements further logic to ensure row and column obeisance, and without noticeably slowing the code down, apply dramatically more consistent, properly dominated solution sets, even for incredibly complex search spaces (in this context, say a 10,000 x 10,000 or larger matrix). This is important because in the realm of metaheuristics, like genetic algorithms, a compromise must be made between the algorithm's design and its outcomes: their stochasticity means they do not guarantee a most optimal solution, especially in complex scenarios like domination.

## IV.    Experimental Results

      The convergence of a genetic algorithm with the wisdom of crowds designed to take into account the concept of spatial domination in addressing MPD exemplifies the potential of hybrid heuristic approaches in computational theory. Results demonstrate the potential of this convergence using a genetic algorithm approach enriched with the wisdom of crowds to optimize the generating of efficient solutions ensuring complete matrix domination.

      The genetic algorithm does not approach the matrices with a brute-force search but instead evaluates the fitness of each potential solution. Iterating through generations, the algorithm is enriched with the wisdom of crowds approach, so more computational power is required than for brute force methods, however with the sample sizes I worked with this did not become an issue.

      While adjusting different parameter settings, such as changing the sizes of the matrices or the number of generations or the mutation rate, the program showed consistency in its display and its efficiency of results. To reach this stage, many problem-solving steps had to be taken as in its early development multiple generations would attempt to break constraints and place multiple dominators in one row or column. As expected, increasing the size of the matrices significantly slows down the speed of the algorithm, though not linearly or exponentially during my tests. Best solutions for smaller matrices over hundreds of generations can be processed in seconds.

## V.    Data

      The data used for this program is data initiated by the program itself by means of randomly generated matrices using permutations to ensure non-repetition. Hard-coded matrices supplied as arrays could be implemented into the program with a short function if there were other TMDP solution sets to compare with; however, I could not find any such datasets as of November 2023.

# VI. Results

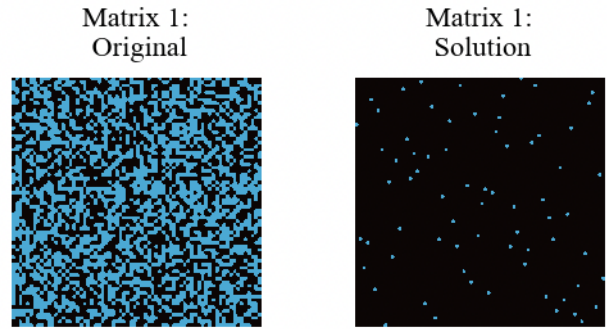

Matrix 1: Original    Matrix 1: Solution

The visualization component from matplotlib, in addition to the terminal display of the placement of 1s and 0s in the matrix, was a reliable way to quickly check how well the algorithm was performing. The results demonstrated that the genetic algorithm enriched with the wisdom of crowds metaheuristic was able to grow and perform much more optimally than other methods, such as brute force or the genetic algorithm without the wisdom of crowds implementation. The size of the matrices for TMDP did not affect the greater efficiency of the wisdom of crowds approach compared to without.

# VII. Conclusions

This research has explored the application of a genetic algorithm metaheuristic enriched by the wisdom of crowds in order to address TMDP. Integrating the exploratory nature of a genetic algorithm with the wisdom of crowds to find more optimal solutions with user-defined parameters to work within computational complexity considerations and gauge performance mainly with a fitness evaluation function and a constraining function to combat the stochastic nature of genetic algorithms, this approach emphasizes collective decision-making in the selection process.

Results demonstrate the potential of this convergence to generate efficient solutions, optimizing the trade-off between the number of dominators and their strategic placements within the matrices while efficiently ensuring consistent and total matrix domination. This approach, combining the metaheuristic of a genetic algorithm with the wisdom of crowds, offers something new when approaching the NP-Complete TMDP and should be applied to some of its sister NP-Complete problems to determine more underlying similarities in the nature of finding optimal solution sets. TMDP and related problems indeed have a myriad of practical, significant reasons for relevance.